\documentclass[10pt,twocolumn,letterpaper]{article}

\usepackage[pagenumbers]{cvpr} 

\usepackage{graphicx}
\usepackage{amsmath}
\usepackage{amssymb}
\usepackage{booktabs, multirow}
 
\usepackage{cuted}  
\usepackage{capt-of}  

\usepackage[pagebackref,breaklinks,colorlinks]{hyperref}

\usepackage[capitalize]{cleveref}
\crefname{section}{Sec.}{Secs.}
\Crefname{section}{Section}{Sections}
\Crefname{table}{Table}{Tables}
\crefname{table}{Tab.}{Tabs.}

\def\cvprPaperID{5967} 
\def\confName{CVPR}
\def\confYear{2022}

\begin{document}

\title{3D Shape Variational Autoencoder Latent Disentanglement \\ via Mini-Batch Feature Swapping for Bodies and Faces}

\author{
    Simone Foti \quad Bongjin Koo \quad Danail Stoyanov \quad Matthew J. Clarkson \\
    University College London \\
    {\tt\small s.foti@cs.ucl.ac.uk}
}
\maketitle

\begin{strip}
    \centering
    \captionsetup{type=figure}
    \includegraphics[width=\textwidth]{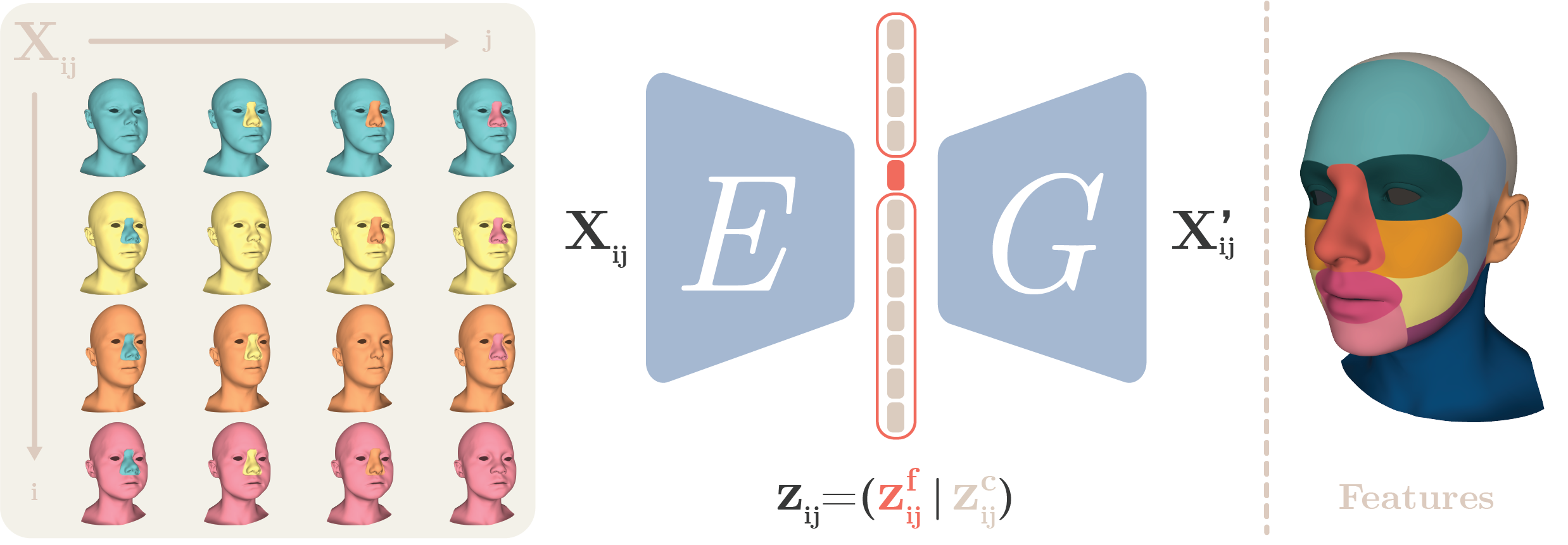}
    \captionof{figure}{Schematic Description of the Proposed Method. \textit{Left}: an arbitrary identity feature is selected and a mini-batch of vertices ($\mathbf{X}_{ij}$) is created by swapping features across different 3D shapes. Colours represent identities. Notice that features from the same identity have the same colour. 
    \textit{Centre}: a 3D-VAE ($\{E, G\}$) encodes $\mathbf{X}_{ij}$ in its latent representations $\mathbf{z}_{ij}=(\mathbf{z}^f_{ij}|\mathbf{z}^c_{ij})$, which are subsequently decoded into $\mathbf{X}'_{ij}$. In this case $f$ corresponds to the nose. Therefore, while $\mathbf{z}^f_{ij}$ controls the shape of the nose, $\mathbf{z}^c_{ij}$ controls the shape of the rest of the face. \textit{Right}: visual representation of all the different mesh features for which we seek to obtain a disentangled latent representation.}
    \label{fig:method}
\end{strip}

\begin{abstract}
   Learning a disentangled, interpretable, and structured latent representation in 3D generative models of faces and bodies is still an open problem. The problem is particularly acute when control over identity features is required. 
   In this paper, we propose an intuitive yet effective self-supervised approach to train a 3D shape variational autoencoder (VAE) which encourages a disentangled latent representation of identity features. Curating the mini-batch generation by swapping arbitrary features across different shapes allows to define a loss function leveraging known differences and similarities in the latent representations. Experimental results conducted on 3D meshes show that state-of-the-art methods for latent disentanglement are not able to disentangle identity features of faces and bodies. Our proposed method properly decouples the generation of such features while maintaining good representation and reconstruction capabilities. 
   Our code and pre-trained models are available at \href{http://www.github.com/simofoti/3DVAE-SwapDisentangled}{github.com/simofoti/3DVAE-SwapDisentangled}.
\end{abstract}


\section{Introduction}
    \label{sec:intro}
    The generation of 3D human faces and bodies is a complex task with multiple potential applications ranging from movie and game productions, to augmented and virtual reality, as well as healthcare applications. Currently, the generation procedure is either manually performed by highly skilled artists or it involves semi-automated avatar design tools. Even though these tools greatly simplify the design process, they are usually limited in flexibility because of the intrinsic constraints of the underlying generative models~\cite{gruber2020interactive}. 
    Blendshapes~\cite{loper2015smpl, osman2020star, tena2011interactive}, 3D morphable models~\cite{blanz1999morphable, ploumpis2019combining, li2017learning}, autoencoders~\cite{ranjan2018generating, gong2019spiralnet++, aumentado2019geometric, cosmo2020limp}, and generative adversarial networks~\cite{cheng2019meshgan, gecer2020synthesizing, li2020learning, abrevaya2019decoupled} are currently the most used generative models, but they all share one particular issue: the creation of local features is difficult or even impossible. In fact, not only do generative coefficients (or latent variables) lack any semantic meaning, but they also create global changes in the output shape.
    For this reason, we focus on the problem of 3D shape creation by enforcing disentanglement among sets of generative coefficients controlling the identity of a character.
    
    Following~\cite{bengio2013representation, higgins2016beta, kim2018disentangling} we define a disentangled latent representation as one where changes in one latent unit affects only one factor of variation while being invariant to changes in other factors. More interpretable and structured latent representations of data that expose their semantic meaning have been widely researched in the artificial intelligence community~\cite{higgins2016beta, kim2018disentangling, kulkarni2015deep, esmaeili2019structured, ding2020guided}, but this is still an open problem especially for generative models of 3D shapes~\cite{aumentado2019geometric}. 
    Given the superior representation capabilities, the reduced number of parameters, and the stable training procedures, we decide to focus our study on deep-learning-based generative models and in particular on variational autoencoders (VAEs).
    In this field, recent work has tried to address the latent disentanglement problem for 3D shapes and managed to decouple the control over identity and expression (or pose)~\cite{aumentado2019geometric, cosmo2020limp, abrevaya2019decoupled}, but they are still unable to properly disentangle identity features. Some success has been achieved in the generation of 3D shapes of furniture~\cite{nash2017shape, yang2020dsm}, but the structural variability of the data requires complex architectures with multiple encoders and decoders for different furniture parts. In contrast, our method relies on a single VAE which is trained by curating the mini-batch generation procedure and with an additional loss. The intuition behind our method is that if we swap features (e.g. nose, ears, legs, arms, etc.) across the input data in a controlled manner (Fig.~\ref{fig:method}, \textit{Left}), we not only know a priori which shapes within a mini-batch have (do not have) the same feature, but we also know which are (are not) created from the same face (body). These differences and similarities across shapes should be captured in the latent representation. Therefore, assuming that different subsets of latent variables correspond to different features, we can partition the latent space and leverage the structure of the input batch to encourage a more disentangled, interpretable, and structured representation. 
    
    With the objective of building a model capable of generating 3D meshes, we define our VAE architecture extending~\cite{gong2019spiralnet++}. This state-of-the-art model proved to be fast and capable of better capturing non-linear representations of 3D meshes, while leveraging very intuitive convolutional operators characterised by a reduced number of parameters. Nonetheless, the network choice is arbitrary and we expect our method to be working also with other network configurations and operators. Even though we consider meshes as our primary data structures, it is also worth noting that, by providing semantic segmentations of the different features, our method is applicable to voxel- or point-cloud-based generative models. We believe that the generality of the proposed method is particularly important in the current geometric deep learning field, where definitions of 3D convolutions and pooling operators are still an open problem. 
    
    To summarise, the key contributions of our approach are:
    (\emph{i}) the definition of a new mini-batching procedure based on feature swapping,
    (\emph{ii}) the introduction of a novel loss function capable of leveraging shape differences and similarities within each mini-batch, and
    (\emph{iii}) the consequent creation of a 3D-VAE capable of generating 3D meshes from a more interpretable and structured latent representation.

\section{Related Work}
    \label{sec:related_work}
    In this section, we first discuss existing work on 3D generative models of faces and bodies, followed by state-of-the art approaches for latent disentanglement of autoencoder-based generative models.
    
    \paragraph{Generative Models}
        Blendshape models manually created by artists linearly interpolate local features between two or more manually selected shapes. These models are common as consumer-level avatar design tools adopted by several videogame engines. Even though they guarantee control over the generation of local features, they are very large models usually built with only a few subjects and are capable of offering only very limited flexibility and expressivity~\cite{gruber2020interactive}.
        A widespread approach to overcome these limitations is to rely on linear statistical 3D morphable models (3DMM). These models are based upon the identity space of a population of 3D shapes, and are usually built by applying a principal component analysis (PCA) over the entire dataset. They are always built with the assumption that shapes are registered between each other and in dense point correspondence. This allows the generation of meaningful and morphologically realistic shapes as linear combinations of training data. This technique was pioneered by \cite{blanz1999morphable} and further developed and adopted by many researchers~\cite{egger20203d}. Interestingly, \cite{gruber2020interactive} divided the face in different local patches and trained a PCA model for each region in order to control the generation of different facial features. The generation of new faces and interactive face sculpting are then achieved through a constrained optimisation. Recently, \cite{ploumpis2019combining, ploumpis2020towards} combined multiple 3DMMs to create the first combined, large-scale, full-head morphable model. In particular, the universal head model (\textsc{Uhm})~\cite{ploumpis2019combining} combines the Large-Scale Face Model (LSFM)~\cite{booth20163d}, which was built with face scans from $10,000$ subjects, with the LYHM head model~\cite{dai2020statistical}. In \cite{ploumpis2020towards} it was extended by combining also a detailed ear model, eye and eye region models, as well as basic models for mouth, teeth, tongue and inner mouth cavity. As further detailed in Sec.~\ref{sec:experiments}, given the high diversity of \textsc{Uhm} we decided to train our face model on heads from \cite{ploumpis2019combining}. 
        
        PCA-based models and blendshapes are often combined. For instance, \textsc{Smpl}\cite{loper2015smpl} learns linear PCA models of male and female body shapes from approximately $2,000$ scans per gender, and subsequently uses the resulting principal components as body shape blendshapes capable of efficiently controlling the identity of a subject. The same approach is used also by \textsc{Star}~\cite{osman2020star}, which not only creates more realistic pose deformations than~\cite{loper2015smpl}, but it also leverages $10,000$ additional scans to improve the generalisation capabilities of the model. Given its better generalisation with respect to other state-of-the-art methods, we trained our body model on shapes generated from \textsc{Star}.
        
        Recently, advances in the geometric deep learning community allowed to efficiently define convolutional operators on 3D data such as meshes and point-clouds. \cite{ranjan2018generating} is the first AE for 3D meshes of faces based on a graph convolutional neural network. This model was built using significantly less parameters than state-of-the-art PCA-based models and showed lower reconstruction errors as well as better generalisation to unseen faces. Other AE-based architectures leveraging different convolutional operators over different datasets were subsequently introduced~\cite{aumentado2019geometric, cosmo2020limp, litany2018deformable, yuan2020mesh, zhou2020fully}. Despite the remarkable performance of these models, we decided to adopt the base architecture of \cite{gong2019spiralnet++}, which further improved upon previous methods by defining a more intuitive convolutional operator based on dilated spiral convolutions (i.e. spiral++ convolution). 
        
        An alternative line of work considers generative adversarial networks (GANs) instead of autoencoders. The first GAN operating on 3D meshes was proposed in~\cite{cheng2019meshgan} and it allowed to disentangle identity from expression generative factors. Other methods usually map 3D shapes to the image domain and then train adversarial networks with traditional 2D convolutions~\cite{abrevaya2019decoupled, gecer2020synthesizing, li2020learning}. 
        GAN models are generally able to generate more detailed and realistic 3D shapes than autoencoders at the cost of being more unstable and difficult to train.
        
        As aforementioned in Sec.~\ref{sec:intro}, with the exception of artistically-created blendshape models and \cite{gruber2020interactive}, none of the other methods here described allow to control local changes during the generation process because their generative coefficients lack any semantic meaning, are not easily interpretable and are not properly disentangled. 
        
    \paragraph{Autoencoder Latent Disentanglement}
        Latent disentanglement for the generation of 3D shapes has been explored mostly in relation to the disentanglement of identity and pose generative factors. \cite{aumentado2019geometric} created a two level architecture combining a point-cloud AE with a VAE where the latent space is successfully partitioned by relying on multiple geometric losses and disentanglement penalties. \cite{cosmo2020limp} achieves similar results by training a point-cloud VAE while controlling the amount of distortion incurring in the construction of the latent space. As mentioned in Sec.~\ref{sec:intro}, these methods are not capable of disentangling generative factors controlling the identity of different subjects. Methods such as~\cite{yang2020dsm, nash2017shape}, on the other hand, are able to control different parts of furniture meshes, but they require complex architectures with multiple encoders and decoders controlling the different parts. Even though part hierarchies have to be considered in the model formulation, differently from faces and bodies, discontinuities between different parts are not a problem when generating furniture.
        
        Research on latent disentanglement of AEs often focuses on the scenario in which only raw observations are available without any supervision about the generative factors, and it is usually performed on images. \cite{higgins2016beta} proposed a simple modification to a VAE~\cite{kingma2013auto}. By increasing the weight of the Kullback–Leibler (KL) divergence, $\beta$-VAE showed better latent disentanglement properties at the expense of a reduced quality of the generated samples. Subsequent work, such as~\cite{kumar2017variational, kim2018disentangling}, tried to overcome this limitation. The DIP-VAE~\cite{kumar2017variational} leverages an additional regularisation term on the expectation of the approximate posterior over observed data. The Factor VAE~\cite{kim2018disentangling} encourages the latent distribution to be factorial, and therefore independent across dimensions, by using a latent discriminator and by adding a total correlation term in the VAE loss function. An interesting approach to encourage latent variables to represent predefined transformations was proposed in~\cite{kulkarni2015deep}, where mini-batches are created combining active and inactive transformations and gradients influencing the latent are modified during backpropagation. However, this method requires synthetic datasets created with known properties that can be used during training to achieve the disentanglement. Recently, \cite{esmaeili2019structured} proposed a VAE in which the objective function is hierarchically decomposed to control the relative levels of statistical independence between groups of variables and for individual variables in the same group. The recursive formulation of the loss introduces additional terms for any variable that has to be disentangled and works only where the factors of variation are uncorrelated scalar variables, a requirement that hampers the applicability of the model in real-world scenarios. Finally, the Guided-VAE~\cite{ding2020guided} in its unsupervised setting leverages a secondary decoder that learns a set of PCA bases that are used to guide the training over simple geometrical shapes. Nevertheless, being the secondary decoder based on a PCA, latent variables suffer the same problems of PCA models.
        
        Among the aforementioned methods for latent disentanglement the DIP-VAE~\cite{kumar2017variational} and Factor VAE~\cite{kim2018disentangling} showed good disentanglement performance also on in-the-wild image datasets while requiring only minor modifications to the VAE formulation. For this reason, we implemented a DIP-VAE and a Factor VAE operating on meshes and compared them against our method.

\section{Method}
    The proposed method (Fig.~\ref{fig:method}) allows us to obtain more interpretable and structured latent representations for self-supervised 3D generative models. This is achieved by training a mesh-convolutional variational autoencoder (Sec.~\ref{sec:network}) with a mini-batch controlled feature swapping procedure and a latent consistency loss (Sec.~\ref{sec:minib-swapping}).
    
    \begin{figure}[t]
            \centering
            \includegraphics[width=\linewidth]{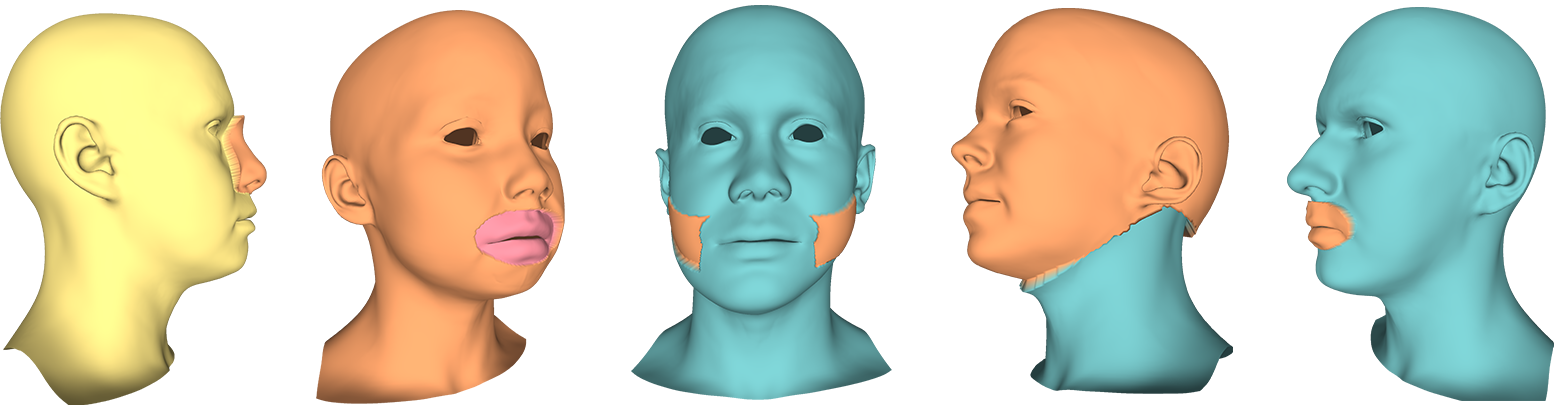}
            \caption{Examples of feature swapping for different features and different subjects.}
            \label{fig:swapped_features}
        \end{figure}
    
    \subsection{Mesh Variational Autoencoder}
        \label{sec:network}
        
        A manifold triangle mesh is defined as $\mathcal{M} = \{\mathbf{X}, \boldsymbol{\mathcal{E}}, \boldsymbol{\mathcal{F}}\}$, where $\mathbf{X} \in \mathbb{R}^{N \times 3}$ is its vertex embedding, $\boldsymbol{\mathcal{E}} \in \mathbb{N}^{\varepsilon \times 2}$ is the edge connectivity that defines its topology, and $\boldsymbol{\mathcal{F}} \in \mathbb{N}^{\Gamma \times 3}$ are its triangular faces. Assuming that meshes share the same topology across the entire dataset, $\boldsymbol{\mathcal{E}}$ and $\boldsymbol{\mathcal{F}}$ are constant and meshes differ from one another only for the position of their vertices, which are assumed to be consistently aligned, scaled, and with point-wise correspondences. 
        Since traditional convolutional operators are not compatible with the non-Euclidean nature of meshes, we build our generative model with the simple yet efficient approach defined in~\cite{gong2019spiralnet++}. 
        Convolution operators are thus defined as learnable functions over pre-computed dilated spiral sequences~\cite{gong2019spiralnet++}. 
        Pooling and un-pooling operators are defined as sparse matrix multiplications with pre-computed transformations that are obtained with a quadric sampling procedure~\cite{gong2019spiralnet++, ranjan2018generating} (see Supplementary Materials).
        
        Our 3D-VAE is built as an encoder-decoder pair (Fig.~\ref{fig:method},~\textit{Centre}), where the decoder is used as a generative model and is referred to as generator. Following this convention, we define our architecture as a pair of non-linear functions $\{E, G\}$. Let $\mathcal{X}$ be the vertex embedding domain and $\mathcal{Z}$ the latent distribution domain, we have $E: \mathcal{X} \rightarrow \mathcal{Z}$ defined as a variational distribution $q(\mathbf{z}|\mathbf{X})$ that approximates the intractable model posterior distribution, and $G: \mathcal{Z} \rightarrow \mathcal{X}$ described by the likelihood $p(\mathbf{X}| \mathbf{z})$.
        Throughout the entire network, each spiral++ convolutional layer is followed by an ELU activation function. However, in $E$ convolutions 
        are interleaved with pooling layers and in $G$ by un-pooling layers. There are also three fully connected layers: two of them are the last layers of $E$ predicting the mean and the diagonal covariance of the variational distribution, the other is the first layer of $G$ and transforms $\mathbf{z} \sim \mathcal{Z}$ back into a low-dimensional mesh that can be processed by mesh convolutions. 
        
        %
        During training, the following loss is minimised:
        \begin{equation}
            \label{eq:vae_loss}
            \mathcal{L}_{VAE} = \mathcal{L}_R + \alpha \mathcal{L}_L + \beta \mathcal{L}_{KL}
        \end{equation}
        \noindent where $\alpha$ and $\beta$ are weighting constants. $\mathcal{L}_R = \frac{1}{N} \sum_{n=1}^{N} \|\mathbf{x}'_{n} - \mathbf{x}_{n} \|_2^2 $ is the mean squared error between the input ($\mathbf{x}_n \in \mathbf{X}$) and the corresponding output ($\mathbf{x}'_n \in \mathbf{X}'=G(E(\mathbf{X}))=G(\mathbf{z})$) vertices. This reconstruction loss encourages the output of the VAE to be as close as possible to its input. $\mathcal{L}_{KL} = KL[q(\mathbf{z}|\mathbf{X})||p(\mathbf{z})]$ is the Kullback–Leibler (KL) divergence pushing the variational distribution towards the prior distribution $p(\mathbf{z})$, which is defined as a standard spherical Gaussian distribution. Finally, $\mathcal{L}_L$ is a smoothing term based on the uniform Laplacian \cite{nealen2006laplacian} that is computed on the output vertices as:
        \begin{equation*}
            \mathcal{L}_L = \frac{1}{N} \sum_{n=1}^{N} \| \mathbf{\delta}_{n} \|_2 \quad \text{with} \; \mathbf{\delta}_{n} = \frac{1}{|\mathcal{N}_n|} \sum_{e \in \mathcal{N}_n} \mathbf{x}'_{e} - \mathbf{x}'_{n} 
        \end{equation*}
        
        \noindent where $\mathbf{\delta}_n$ is the Laplacian of the n-th output vertex, and $\mathcal{N}_n$ the set of its neighbouring vertices with cardinality $|\mathcal{N}_n|$. 
        $\mathcal{L}_L$ is efficiently computed by relying on matrix operators. Concretely, we have $\mathbf{\Delta} = [\mathbf{\delta}_{1},...,\mathbf{\delta}_{N}]^T = \mathbf{L} \mathbf{X}'$, where $\mathbf{L}=\mathbf{I} - \mathbf{D}^{-1} \mathbf{A}$ is the Laplacian operator with random walk normalisation, $\mathbf{A} \in \mathbb{N}^{N \times N}$ is the adjacency matrix and $\mathbf{D} \in \mathbb{R}^{N \times N}$ is the diagonal degree matrix with $D_{aa} = \sum_a A_{ab}$. 
        Note that vertices are normalised by subtracting the per-vertex mean of the training set and dividing the result by the per-vertex standard deviation of the training set, thus the losses in Eq.~\ref{eq:vae_loss} are computed on normalised vertices.
        Also, all loss terms are reduced across mini-batches with a mean reduction.
    
        \begin{figure*}[t]
            \centering
            \includegraphics[width=\textwidth]{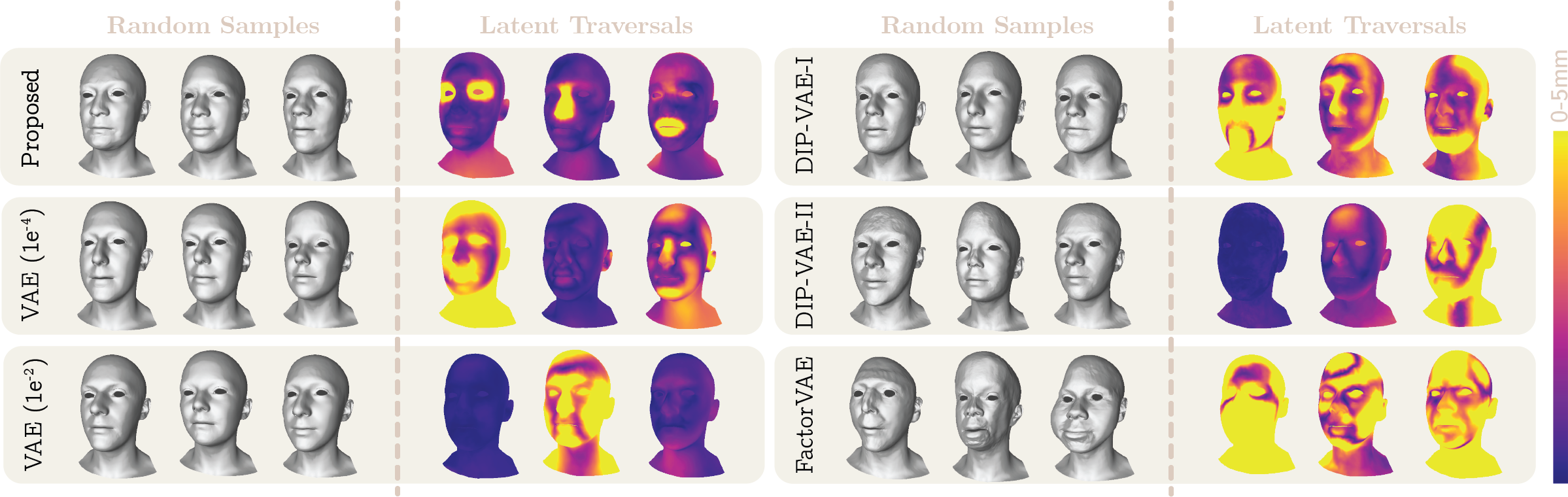}
            \caption{Random samples and vertex-wise distances showing the effects of traversing three randomly selected latent variables (see Supplementary Material to observe the effects for all the latent variables.)}
            \label{fig:rnd_z_comp}
        \end{figure*}
    
    \subsection{Mini-Batch Feature Swapping and Latent Consistency Loss}
        \label{sec:minib-swapping}
        
        We aim to obtain a generative model where vertices corresponding to specific mesh features are controlled by a predefined set of latent variables. Therefore, we start by defining $F$ arbitrary mesh features on a mesh template (Fig.~\ref{fig:method},~\textit{Right}). Features are manually defined by colouring mesh vertices. Since vertices have point-wise correspondences (Sec.~\ref{sec:network}), features can be easily identified for every other mesh in the dataset without manually segmenting them.
        This allows us to swap features from one mesh to another by replacing the vertices corresponding to the selected feature (Fig.~\ref{fig:swapped_features}). 
        
        Feature swapping is at the core of our method and it allows us to curate the mini-batch generation in order to properly shape and constrain the latent representation of each mesh. Each mini-batch of size $B$ can be thought of as a squared matrix of size $\sqrt{B}\times\sqrt{B}$, where each element $\mathbf{X}_{ij}$ is the vertex embedding of a different mesh. As it can be seen from Fig.~\ref{fig:method}~(\textit{Left}), while elements on the diagonal of this matrix are loaded from the dataset, the remaining elements are created online by swapping features. Every time a mini-batch is created, a feature is randomly selected and swapped. Therefore, each row of the matrix contains the same mesh with different features, while each column contains different meshes with the same feature.
        Interestingly, the naive implementation of the feature swapping causes visible surface discontinuities in most input meshes (Fig.~\ref{fig:swapped_features}), but discontinuities are not present in reconstructed meshes thanks to the Laplacian regulariser in Eq.~\ref{eq:vae_loss}. 

        Obviously, when a mini-batch is encoded we obtain a batched latent. As we can see in Fig.~\ref{fig:method}~(\textit{Centre}), for each $\mathbf{X}_{ij}$ we have a corresponding $\mathbf{z}_{ij} \sim E(\mathbf{X}_{ij})$ which is evenly split in $F$ subsets of latent variables, one for each mesh feature ($\mathbf{z}_{ij}=\{\mathbf{z}^{\omega}_{ij}\}_{\omega=1}^{F}$). Note that even though every latent subset $\mathbf{z}^{\omega}_{ij}$ has the same number of variables, uneven splits are also admissible. 
        
        Every time a mini-batch is created by swapping a feature $f$, we can define $\mathbf{z}_{ij} = (\mathbf{z}^{f}_{ij}|\mathbf{z}^{c}_{ij})$. $\mathbf{z}^{f}_{ij}$ is the subset of latent variables controlling the feature swapped across the current mini-batch. $\mathbf{z}^{c}_{ij}$ is the part that controls everything else and is defined as $\mathbf{z}^{c}_{ij} = \{\mathbf{z}^{\omega}_{ij}\}_{\omega=1}^{F} \setminus \{\mathbf{z}^{f}_{ij}\}$.  
        Inspired by both triplet losses and \cite{sanyal2019learning}, and thanks to our curated mini-batching, we can enforce differences and similarities in the latent representation of the different $\mathbf{X}_{ij}$ by requiring matched $\mathbf{z}^{\omega}_{ij}$ pairs to have a distance in latent space that is smaller by a margin, $\eta$, than the distance for unmatched pairs. We traverse the diagonal of the mini-batch latent matrix and compare all the elements on the row containing the diagonal element $\mathbf{z}_{ss}$ with those in the column containing $\mathbf{z}_{ss}$ ($\forall s \in \{1,..., \sqrt{B}\}$). When considering $\mathbf{z}^{f}_{ij}$ we enforce latent similarities across columns and latent differences across rows by evaluating: $\|\mathbf{z}^{f}_{ps} - \mathbf{z}^{f}_{qs}\|_2^2 + \eta_1 \leq \|\mathbf{z}^{f}_{sp} - \mathbf{z}^{f}_{sq}\|_2^2$, $\forall s,p,q \in \{1,..., \sqrt{B}\}$ with $p \neq q$.
        This is justified by the fact that elements in $\mathbf{X}_{ij}$ have the same mesh feature across columns and different mesh features across rows. Vice versa, when considering $\mathbf{z}^{c}_{ij}$, which controls all the other mesh features for the current mini-batch, we enforce similarities row-wise and differences column-wise by evaluating:
        $\|\mathbf{z}^{c}_{sp} - \mathbf{z}^{c}_{sq}\|_2^2 + \eta_2 \leq \|\mathbf{z}^{c}_{ps} - \mathbf{z}^{c}_{qs}\|_2^2,$ $\forall s,p,q \in \{1,..., \sqrt{B}\}$ with $p \neq q$.
        We thus define our latent consistency loss as:

        \begin{equation}
            \label{eq:z_cons_loss}
            \begin{split}
                \mathcal{L}_{c} & = \gamma \hspace{-0.5em} \sum^{\sqrt{B}}_{\substack{s,p,q=1 \\ p \neq q}} \hspace{-0.3em} \max\Big[0, \|\mathbf{z}^{f}_{ps} - \mathbf{z}^{f}_{qs}\|_2^2 - \|\mathbf{z}^{f}_{sp} - \mathbf{z}^{f}_{sq}\|_2^2 + \eta_1\Big] + \\ 
                & \qquad + \max\Big[0, \|\mathbf{z}^{c}_{sp} - \mathbf{z}^{c}_{sq}\|_2^2 - \|\mathbf{z}^{c}_{ps} - \mathbf{z}^{c}_{qs}\|_2^2 + \eta_2\Big]
            \end{split}
        \end{equation}
        
        \noindent where $\gamma = \frac{1}{B\sqrt{B} - B}$ is a batch normalisation term that considers all the latent distances comparisons performed while computing $\mathcal{L}_{c}$. Combining Eq.~\ref{eq:vae_loss} with Eq.~\ref{eq:z_cons_loss} and said $\kappa \in \mathbb{R}$ a weighting coefficient, we can formulate the total loss as: 
        
        \begin{equation}
            \label{eq:tot_loss}
            \mathcal{L} = \mathcal{L}_{VAE} + \kappa \mathcal{L}_{c} = \mathcal{L}_{R} + \alpha \mathcal{L}_{L} + \beta \mathcal{L}_{KL} + \kappa \mathcal{L}_{c}
        \end{equation}

\section{Experiments}
    \label{sec:experiments}
    
    \begin{figure*}
            \centering
            \includegraphics[width=\textwidth]{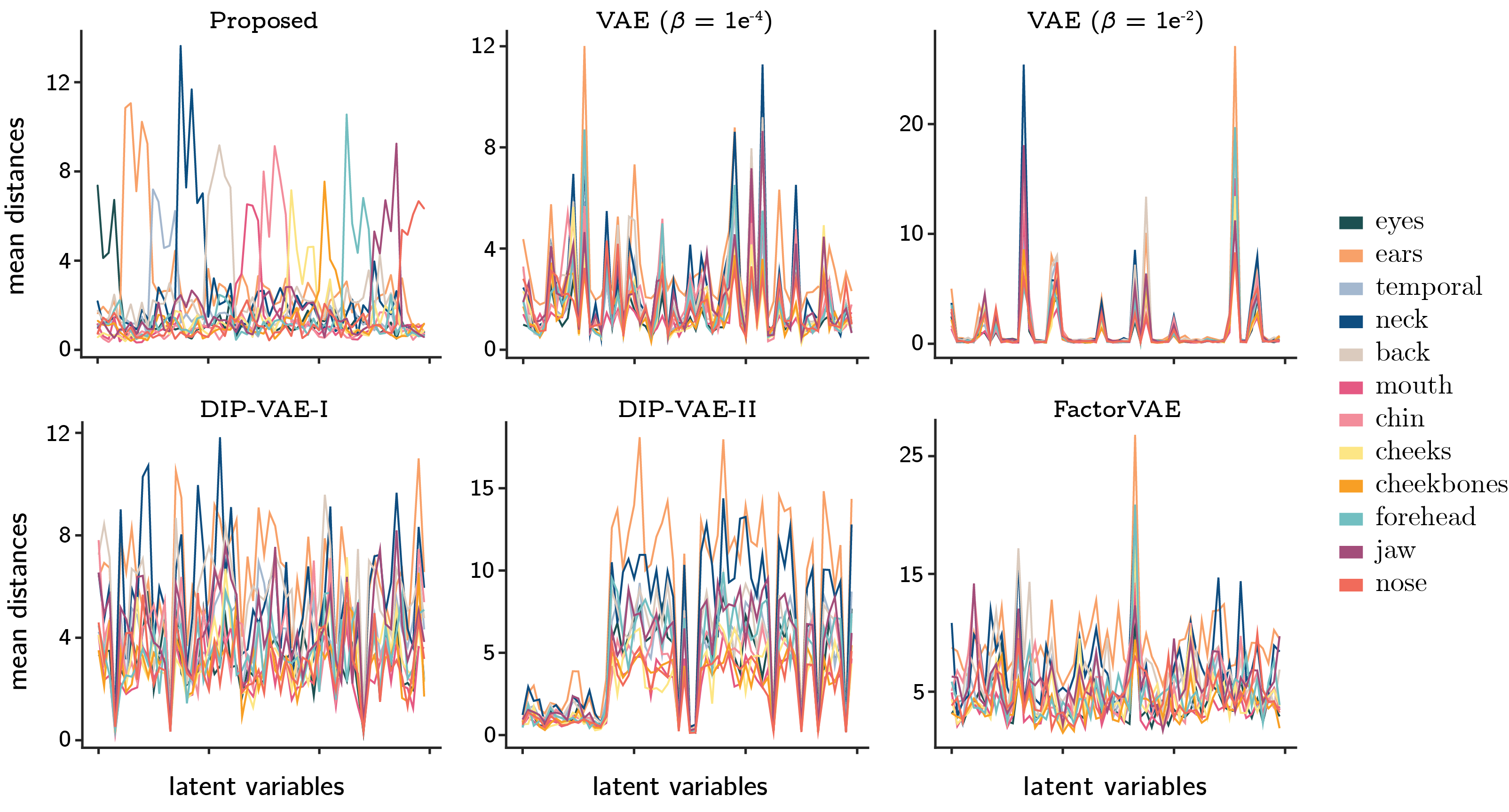}
            \caption{Effects of traversing each latent variable across different mesh features. For each latent variable (abscissas) we represent the per-feature mean distances computed after traversing the latent variable from its minimum to its maximum value. 
            For each latent variable, we expect a high mean distance in one single feature and low values for all the others.}
            \label{fig:z_compared_cond}
        \end{figure*}
    
    \paragraph{Datasets}
        Our main objective is to train a generative model capable of generating different identities from a set of feature-disentangled latent variables. For our experiments we require datasets containing as many subjects as possible in a neutral expression. However, most open source datasets for 3D shapes of faces, bodies, or animals contain only a limited number of subjects captured in different expressions or poses (e.g. \textsc{Mpi}-Dyna~\cite{pons2015dyna}, \textsc{Smpl}~\cite{loper2015smpl}, \textsc{Surreal}~\cite{varol2017learning}, Co\textsc{ma}~\cite{ranjan2018generating}, \textsc{Smal}~\cite{zuffi20173d}, etc.). For this reason, we rely on two linear models that were built using a conspicuous number of subjects and that are released for non-commercial scientific research purposes: \textsc{Uhm}~\cite{ploumpis2019combining} and \textsc{Star}~\cite{osman2020star} (Sec.~\ref{sec:related_work}). From these models we randomly generate $10,000$ meshes and create one dataset for faces and one for bodies. We use $90\%$ of the data for training, $5\%$ for validation, and $5\%$ for testing.
    
    \paragraph{Implementation Details}
        All networks were implemented in PyTorch and trained for $40$ epochs using the \textsc{Adam} optimiser~\cite{kingma2014adam} with a fixed learning rate of $lr=1e^{-4}$ and mini-batch size $B=16$
        (note that the feature swapping is applied to our method only).
        Spiral convolutions~\footnote{The SpiralNet++ implementation was made available with an MIT license.} had spiral length of $9$ and spiral dilation of $1$. The last convolutional layer of $E$ and the first of $G$ had $64$ features while all the others $32$. The sampling factors used during the quadric sampling for the creation of the up- and down-sampling transformation matrices were set to $4$. 
        Since the two datasets have a significantly different number of vertices ($N_{faces}=71,928$ and $N_{bodies}=6,890$), networks operating on faces have $4$ convolutional layers interleaved with sampling operators in both $E$ and $G$, while networks operating on bodies have only $3$. For the same reason latent sizes are different: $60$ variables for faces and $33$ for bodies. Considering that the face template was segmented in $12$ regions and the body template in $11$, each $\mathbf{z}^{\omega}_{ij}$ has $5$ variables for faces and $3$ for bodies. The weight of the Laplacian regulariser was set to $\alpha=1$, 
        while the latent consistency weight was $\kappa=0.5$ for faces and $\kappa=1$ for bodies. $\eta_1$ and $\eta_2$ were set to $\eta_1=\eta_2=0.5$ for both datasets. Training was performed on a single Nvidia Quadro P5000 for faces and on an Nvidia GeForce GTX 1050Ti for bodies. We run approximately $120$ experiments in $25$ GPU days.

    \paragraph{Comparison with Other Methods}
        \begin{table*}
            \caption{Quantitative comparison between our model and other state-of-the-art methods for self-supervised latent disentanglement. All methods were trained on the same face dataset. 
            Mean and Max Rec. refer to the the mean and maximum average per-vertex reconstruction errors across the test set. Values are computed in millimetres. The diversity is computed as detailed in Sec.~\ref{sec:experiments}. All the other metrics for the evaluation of the generation capabilities were introduced in~\cite{yang2019pointflow}.}
            \label{tab:recon-div}
            \centering
            \begin{tabular}{l cccc cc cc cc}
                \toprule
                    \multirow{2}{*}{Method} & \multirow{2}{*}{\parbox[c]{.07\linewidth}{\centering Mean Rec. ($\downarrow$)}} & \multirow{2}{*}{\parbox[c]{.07\linewidth}{\centering Max Rec. ($\downarrow$)}} & \multirow{2}{*}{\parbox[c]{.08\linewidth}{\centering Diversity ($\uparrow$)}} & \multirow{2}{*}{JSD ($\downarrow$)} & \multicolumn{2}{c}{MMD ($\downarrow$)} & \multicolumn{2}{c}{COV({\footnotesize \%}, $\uparrow$)} & \multicolumn{2}{c}{1-NNA ({\footnotesize $\Delta$\%}, $\downarrow$)} \\
                    \cmidrule(lr){6-7} \cmidrule(lr){8-9} \cmidrule(lr){10-11}
                    & & & & & CD & EMD & CD & EMD & CD & EMD \\ 
                \midrule
                    VAE ($\beta=1e^{-2})$  & $1.47$     & $1.99$    & $5.43$  & $\mathbf{1.55}$  & $1.66$    & $0.43$     & $62.99$ & $63.67$ & $7.25$         & $7.50$          \\
                    VAE ($\beta=1e^{-4})$  & $\mathbf{0.61}$     & $\mathbf{0.74}$    & $4.23$  & $4.89$  & $1.53$    & $0.38$     & $65.49$ & $\mathbf{66.33}$ & $1.17$         & $\mathbf{0.17}$          \\
                    DIP-VAE-I              & $4.65$     & $11.86$   & $4.74$  & $5.32$  & $\mathbf{1.24}$    & $\mathbf{0.29}$     & $55.57$ & $56.42$ & $4.31$         & $4.56$          \\
                    DIP-VAE-II             & $4.76$     & $11.92$   & $4.30$  & $6.44$  & $1.70$    & $0.43$     & $48.48$ & $47.30$ & $17.72$        & $17.15$         \\
                    Factor VAE             & $0.74$     & $1.01$    & $\mathbf{10.51}$ & $12.47$ & $3.60$    & $0.97$     & $41.05$ & $41.05$ & $2.28$         & $2.62$          \\
                    Proposed               & $0.73$     & $0.93$    & $4.23$  & $4.30$  & $1.56$    & $0.38$     & $\mathbf{65.67}$ & $63.67$ & $\mathbf{0.50}$         & $1.67$          \\
                \bottomrule
            \end{tabular}
        \end{table*}
        
        We compare our method with other self-supervised methods based on encoder-decoder pairs. For a fair comparison, all methods share the same underlying architecture, which we refer to as VAE and which is already detailed in Sec.~\ref{sec:network}. Consistently with the current literature~\cite{ranjan2018generating, litany2018deformable, foti2020intraoperative, yuan2020mesh}, we found that the weight coefficient ($\beta$) on the  KL divergence in VAEs for meshes is smaller than the one used for images. In fact, with $\beta \geq 1$ the VAE is not able to reconstruct the data. Thus, we report results on VAEs with $\beta \in \{1e^{-2}, 1e^{-4}\}$. It is worth noting that the discrepancy between meshes and images does not allow to define a $\beta$-VAE with the same criteria used in the literature ($\beta > 1$)~\cite{higgins2016beta}. We also compare our method with the DIP-VAE-I, DIP-VAE-II, and Factor VAE. To the best of our knowledge this is the first attempt to use them in the mesh domain. Therefore, for the two DIP-VAEs, we set $\beta=1e^{-4}$ and, following the hyperparameter tuning strategy adopted in the original implementation~\cite{kumar2017variational}, we tune $\lambda_d$ and $\lambda_{od}$. Here we report results for DIP-VAE-I with $\lambda_d=100$ and $\lambda_{od}=10$ as well as for DIP-VAE-II with $\lambda_d=10$ and $\lambda_{od}=10$, which qualitatively showed better disentanglement performances. Factor VAE is trained with a discriminator learning rate of $1e^{-6}$, and a total correlation weight $\gamma=0.25$.
        
        We first evaluate the quality of the different models trained on the face dataset in terms of reconstruction errors, diversity of the generated samples, Jensen-Shannon Divergence (JSD)~\cite{achlioptas2018learning}, Coverage (COV)~\cite{achlioptas2018learning}, Minimum matching distance (MMD)~\cite{achlioptas2018learning}, and 1-nearest neighbour accuracy (1-NNA)~\cite{yang2019pointflow} (Tab.~\ref{tab:recon-div}). Mean and maximum reconstruction errors are computed with respect to mean per-vertex errors across the test set. The diversity is computed as the mean of mean per-vertex distances among pairs of meshes randomly generated with the model. The other metrics are computed by leveraging the Chamfer (CD) and Earth Mover (EMD) distances on $2048$ randomly selected pairs of vertices. Note that since the original formulation of 1-NNA expects scores converging to $50\%$, in Tab.~\ref{tab:recon-div} we report absolute differences between the original score and the $50\%$ target value.
        From Tab.~\ref{tab:recon-div} we observe that while most methods for latent disentanglement have significantly increased reconstruction errors, our method closely match the VAE. We also notice that while most models have similar diversity, Factor VAE is able to generate more diverse data. While this property seems to be desirable, observing some randomly generated sample (Fig.~\ref{fig:rnd_z_comp}), we argue that sampled faces are less realistic. The other metrics used to evaluate the generation capabilities of the different models show that our method is comparable with the others, thus proving that our mini-batching procedure and latent consistency loss do not negatively affect the generation capabilities. 
        
        \begin{figure*}
            \centering
            \includegraphics[width=\textwidth]{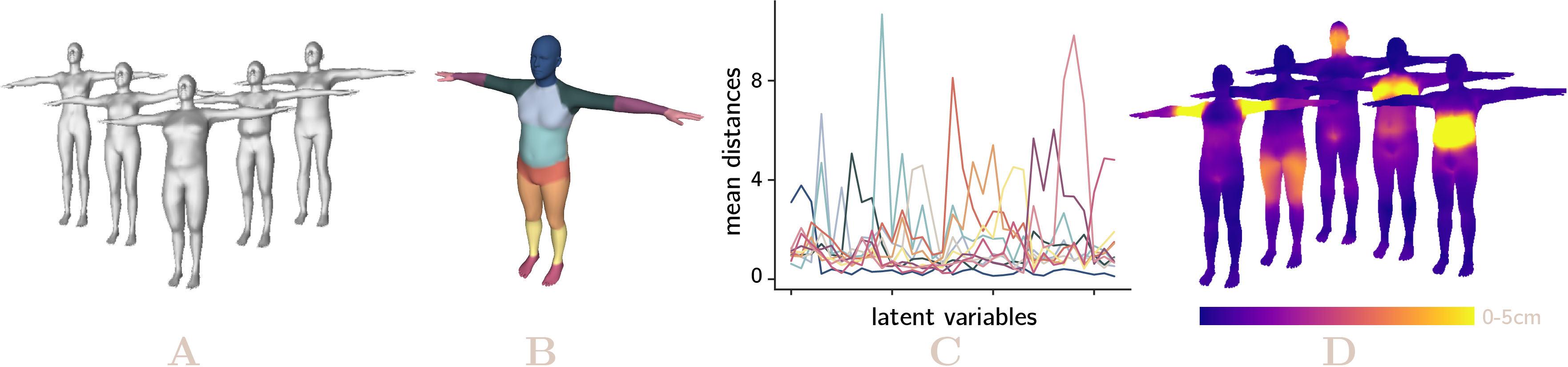}
            \caption{Results of our method on bodies. A: samples randomly generated with the proposed method trained on body meshes. B: visual representation of all the different body features for which we seek to obtain a disentangled latent representation. C: effects of latent variable traversals for each latent variable across different body features. D: vertex-wise distances showing the effects of traversing five latent variables (see Supplementary Material for all the latent variables). }
            \label{fig:body_results}
        \end{figure*}
        
    \paragraph{Evaluation of Latent Disentanglement}
        Previous work evaluated the latent disentanglement on either datasets where labelled data were available or on custom-made datasets of images whose generative factors could be used as labels. Examples of such datasets~\cite{higgins2016beta, kumar2017variational, kim2018disentangling} are binary images of geometric shapes (e.g. circles, rectangles, etc.) where shape deformation parameters are known, or images rendered with controlled camera and lighting positions. Even though both our datasets are generated from existing models, these models lack control over the generative factors, thus traditional metrics such as Z-Diff~\cite{higgins2016beta}, SAP~\cite{kumar2017variational}, and Factor~\cite{kim2018disentangling} scores cannot be computed. In addition, the few unsupervised disentanglement metrics currently existing~\cite{zaidi2020measuring}, are not suitable for our evaluation because \cite{liu2020metrics} is tailored for the evaluation of the disentanglement of style and content information, while \cite{duan2019unsupervised} is used for model and hyperparameter selection thus requiring multiple computationally expensive hyperparameter sweeps. Therefore, we decide to evaluate the effects caused on the generated meshes while traversing each latent variable. We generate two meshes corresponding to each latent variable: one is created setting one latent variable to its minimum ($-3$) and all the remaining to their mean value ($0$), the other replacing the minimum with the maximum value ($+3$). The per-vertex Euclidean distances between the two shapes represent the effects of perturbing a single latent variable. These effects can be qualitatively assessed by observing meshes rendered with vertex colours proportional to the distances (Fig.~\ref{fig:rnd_z_comp}, and Fig.~\ref{fig:body_results}~D). Alternatively, distances corresponding to each feature (Fig.~\ref{fig:method}, \textit{Right} and Fig.~\ref{fig:body_results}~A) can be averaged and subsequently plotted as in Fig.~\ref{fig:z_compared_cond} and Fig.~\ref{fig:body_results}~C. This representation clearly highlights how perturbing each latent variable affects the different features. While most methods appear to be difficult to interpret and mostly entangled, our method shows a significantly more structured, interpretable, and disentangled latent representation than other methods. Interestingly, in the VAE with $\beta=1e^{-2}$ we observe a polarised regime in which only a subset of latent variables control the generated shapes. However, these variables are also controlling the same features, thus disentanglement is not achieved. Since the polarised regime occurs in $\beta$-VAEs~\cite{rolinek2019variational}, we can consider this VAE to be a $\beta$-VAE operating on meshes.

\begin{figure}[t]
            \centering
            \includegraphics[width=\linewidth]{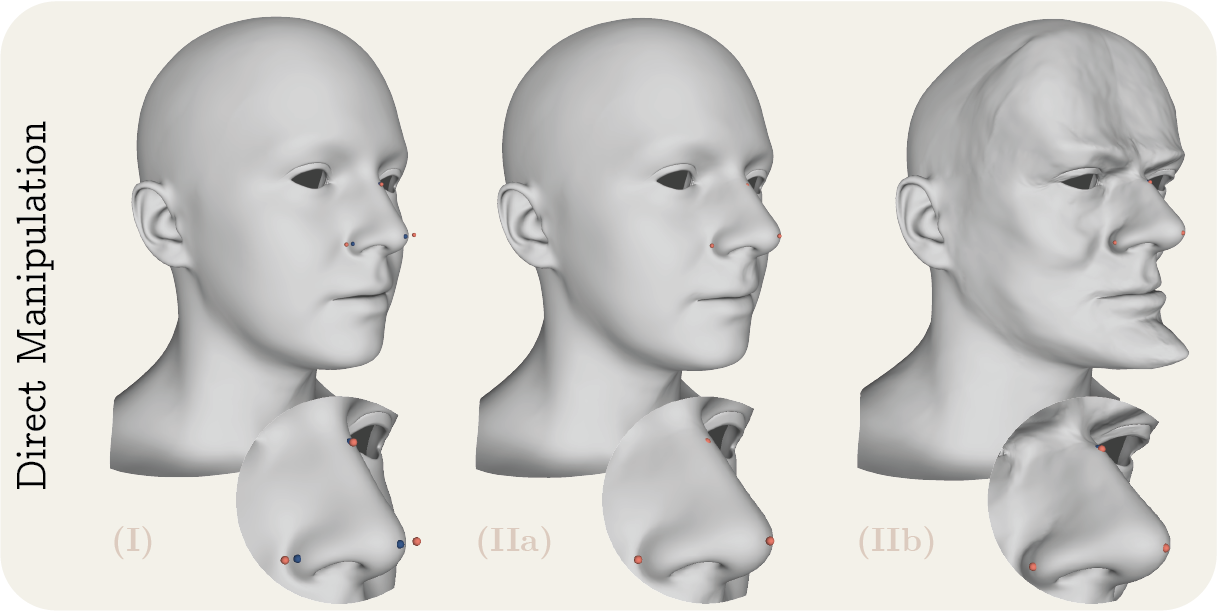}
            \caption{Direct manipulation of the generated mesh. (I) The user selects an arbitrary number of vertices (blue) and their new desired position (red), then our method generates a locally edited mesh fitting the desired locations. Results are reported optimising only $\mathbf{z}^f$ (IIa) and optimising the entire $\mathbf{z}$ (IIb).}
            \label{fig:manip_and_fitting}
        \end{figure}

    \paragraph{Direct Manipulation}
        Similarly to~\cite{gruber2020interactive}, our method supports direct manipulation of the generated 3D meshes. A user is thus able to select one or multiple vertices, specify their new desired location, and our method automatically generates a new mesh locally deformed to satisfy the user edit. This is achieved through a small optimisation procedure over the latent representation. We use the \textsc{Adam} optimiser for $50$ iterations and with a fixed learning rate of $lr=0.1$. Given  $ S \circ \mathbf{X}' = S \circ G(\mathbf{z}) \in \mathbb{R}^{\Upsilon \times 3}$ the subset of vertices manually selected from the currently generated mesh, and their desired positions $\mathbf{Y} \in \mathbb{R}^{\Upsilon \times 3}$, with $\Upsilon$ representing the number of selected vertices, we optimise: $\min_{\mathbf{z}^f} \| S \circ G(\mathbf{z}) - \mathbf{Y}\|_2^2$. Note that the optimisation over $\mathbf{z}^f$ guarantees the locality of the manipulation (Fig.~\ref{fig:manip_and_fitting}, IIa) and it is achieved by setting to zero the gradients computed over $\mathbf{z}^c$. This is made possible by our method and its improved latent disentanglement. An optimisation over the entire latent representation would cause visible global changes (Fig.~\ref{fig:manip_and_fitting}, IIb), thus making impossible the direct manipulation.
        
\section{Conclusion}
    \label{sec:conclusion}
    We proposed a novel approach to learn a more disentangled, interpretable, and structured latent representation for 3D VAEs. This is achieved by curating the mini-batching procedure with feature swapping and introducing an additional latent consistency loss. Even though our method is able to disentangle predefined subsets of latent variables, we do not guarantee orthogonality and disentanglement among the variables within each subsets. Nonetheless, we can increase the number of subsets to achieve finer control over the generated model. The main limitations of our work are the assumptions made on the training data. A consistent scaling and alignment, as well as dense-point correspondences, and a fixed mesh topology are common for generative models of 3D faces (and bodies) and useful for an efficient feature swapping. However, this assumption could be relaxed to make our method suitable for more general 3D problems if a different architecture was implemented and semantic segmentations of each 3D shape were available. 
    Good semantic segmentations are not trivial to obtain for raw data, but methods such as \cite{wang2019dynamic, verma2018feastnet} could be used.
    As future work, we aim at introducing and properly disentangling expressions (or poses) while retaining the superior latent disentanglement over identity features made possible by our method. 
    
\paragraph{Acknowledgement}
    This work was supported by the Wellcome Trust/EPSRC [203145Z/16/Z]. The views expressed in this publication are those of the author(s) and not necessarily those of the Wellcome Trust.

{\small
\bibliographystyle{ieee_fullname}
\bibliography{bibliography}
}

\end{document}


\title{\textit{Supplementary Material} \\ 3D Shape Variational Autoencoder Latent Disentanglement \\ via Mini-Batch Feature Swapping for Bodies and Faces}

\author{
    Simone Foti \quad Bongjin Koo \quad Danail Stoyanov \quad Matthew J. Clarkson \\
    University College London \\
    {\tt\small s.foti@cs.ucl.ac.uk}
}
\maketitle

\section{Mesh Operators}
    Since traditional neural network operators are not well suited for the non-Euclidean nature of meshes, we rely on spiral++ convolutions [16] and on the sampling operators defined in [35].
    
    The creation of spiral sequences is at the core of the adopted convolution. Spirals are a simple yet effective approach to aggregate neighbouring mesh vertices into ordered sequences. Given a vertex, the spiral sequence is obtained by arbitrarily selecting one neighbour and following a clockwise spiral until the spiral length is reached. The receptive field of these convolutional operators can be expanded dilating the spirals (i.e. not selecting certain vertices along the sequence). Denoting by $\mathcal{S}(n, l)$ the spiral centred at vertex $n$ with length $l$, the convolution at layer $k$ is defined as:
    
    \begin{equation*}
        \mathbf{x}_{n}^{(k)} = \text{MLP}^{(k)} \Big(\underset{j \in \mathcal{S}(n, l)}{\mathbin\Vert} \mathbf{x}_{j}^{(k-1)} \Big)
    \end{equation*}
    
    \noindent where $\mathbin\Vert$ is the concatenation operation over the vertices in the spiral $\mathcal{S}(n, l)$, $\mathbf{x}_{n}^{(k)}$ are the vertex features at layer $k$, and MLP is a multilayer perceptron. Note that spirals are fixed during training because they are pre-computed only once for all vertices.
    
    Pooling and un-pooling operators are matrix multiplications between the vertex features of a given layer and a sparse matrix. The sparse matrices are both pre-computed during a mesh simplification procedure that iteratively contracts the two vertices with the smallest quadric error. In particular, the pooling matrix $Q_d \in \{0, 1\}^{N_{k + 1} \times N_k}$ is a sparse matrix where $Q_d(p, q)=1$ if vertex $q$ has been preserved during quadric sampling and $Q_d(p, q)=0$ otherwise. The un-pooling matrix $Q_u \in \mathbb{R}^{N_k \times N_{k + 1}}$ leaves the preserved vertices unchanged by setting $Q_u(q, p)=1$. 
    Contracted vertices are expressed in barycentric coordinates with respect to the closest preserved triangle, and then their corresponding elements in $Q_u$ are set to the barycentric weights. This allows to restore the contracted vertices.
    
\section{Latent Space Interpolation}
\begin{figure*}
        \centering
        \includegraphics[width=\textwidth]{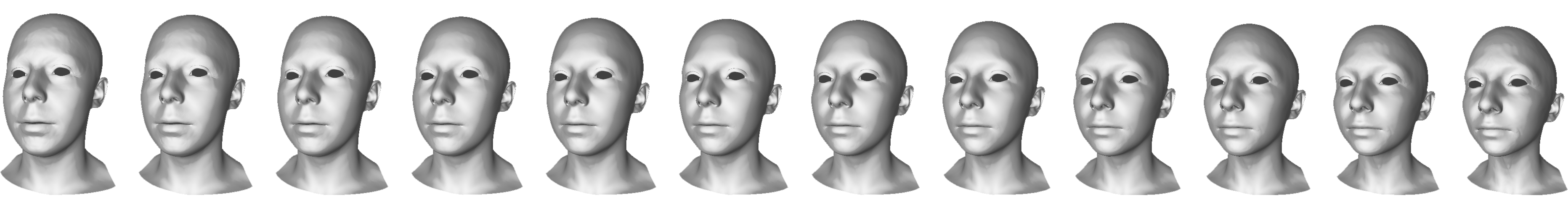}
        \caption{Latent interpolation experiment. An initial and a target shape are selected from the test set. Then, their latent representation $\mathbf{z}$ is computed by feeding the shapes in the encoder network $E$. $10$ intermediate latent vectors are thus computed by linearly interpolating all the latent variables. The shapes generated from these latent vectors smoothly transition from the initial (leftmost shape) to the final (rightmost shape) shape}.
        \label{fig:interpolate}
    \end{figure*}

\begin{figure*}
        \centering
        \includegraphics[width=\textwidth]{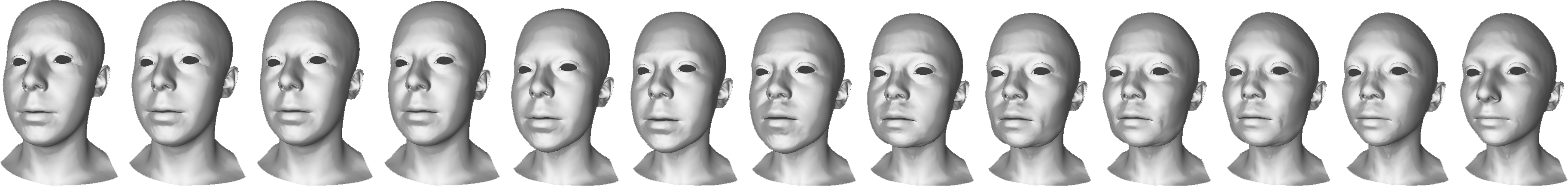}
        \caption{Per-feature latent interpolation experiment. Given the same initial and target latent vectors used in Fig.~\ref{fig:interpolate}, subsets of the latent representation corresponding to different features ($\mathbf{z}^{\omega}$) are progressively replaced. In fact, the first face is the initial face, the second face is obtained by replacing the values of the $\mathbf{z}^{\omega}$ controlling the eye region of the initial face with those controlling the eye region in the target shape. The third face is obtained from the second face by replacing the $\mathbf{z}^{\omega}$ controlling the ears. Then, the subsequent shapes are obtained replacing the $\mathbf{z}^{\omega}$ controlling: temporal, neck, back, mouth, chin, cheeks, cheekbones, forehead, jaw, and nose. Each shape is obtained starting from the one on its left, and therefore the last one also corresponds to the target shape.}
        \label{fig:interpolate_per_feature}
    \end{figure*}

    We performed two latent interpolation experiments. Fig.~\ref{fig:interpolate} shows the effect of interpolating $\mathbf{z}$ between the latent representation of two different shapes. Fig.~\ref{fig:interpolate_per_feature} shows the effects of changing each $\mathbf{z}^{\omega}$ of one shape with the corresponding $\mathbf{z}^{\omega}$ of the other shape, which is equivalent to progressively replacing features of the initial mesh with those of the target mesh. The interpolation experiment of Fig.~\ref{fig:interpolate_per_feature} is better represented in the \textit{supplementary video}~\footnote{The supplementary video is available at the following link \url{https://youtu.be/w9WF0mZe1ig}}. The video also shows the interpolation between each pair of $\mathbf{z}^{\omega}$. The two experiments prove that our method creates a smooth latent space where per-feature modifications are possible. 
    
    The \textit{supplementary video} also shows per-variable latent interpolation experiments for all different methods. Interestingly, while intermediate faces generated with our method are a plausible interpolations between the initial and target shape, intermediate faces generated with other methods often belong to substantially different identities.
    
\section{Random Generation and Latent Disentanglement}
\begin{figure*}
        \centering
        \includegraphics[height=.9\textheight]{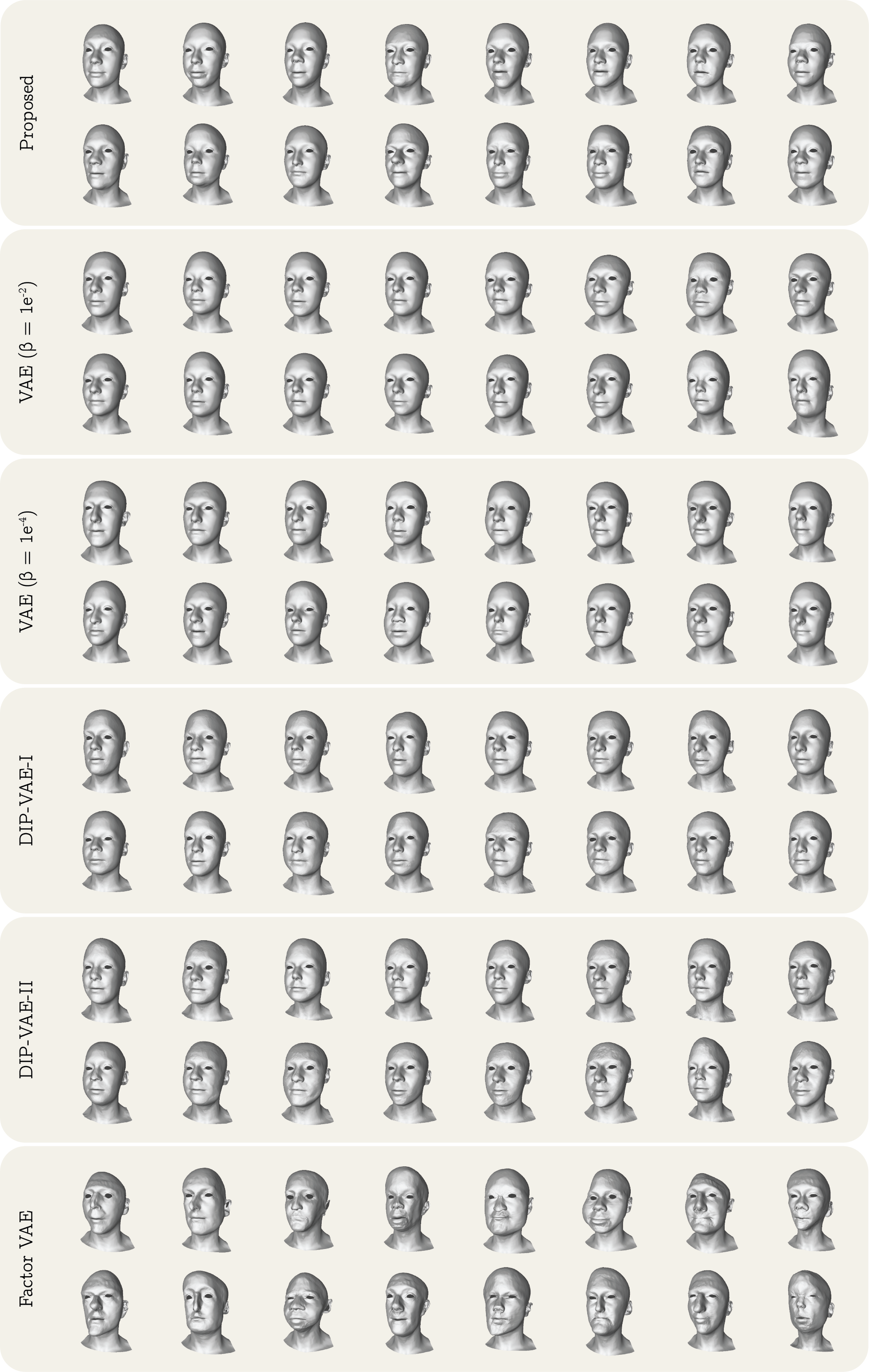}
        \caption{Random sample generation for VAE ($\beta = 1e^{-2}$), VAE ($\beta = 1e^{-4}$), DIP-VAE-I, DIP-VAE-II, and Factor VAE.}
        \label{fig:qualitative_comparisons_rnd_all}
    \end{figure*}

\begin{figure*}[b]
        \centering
        \includegraphics[width=\textwidth]{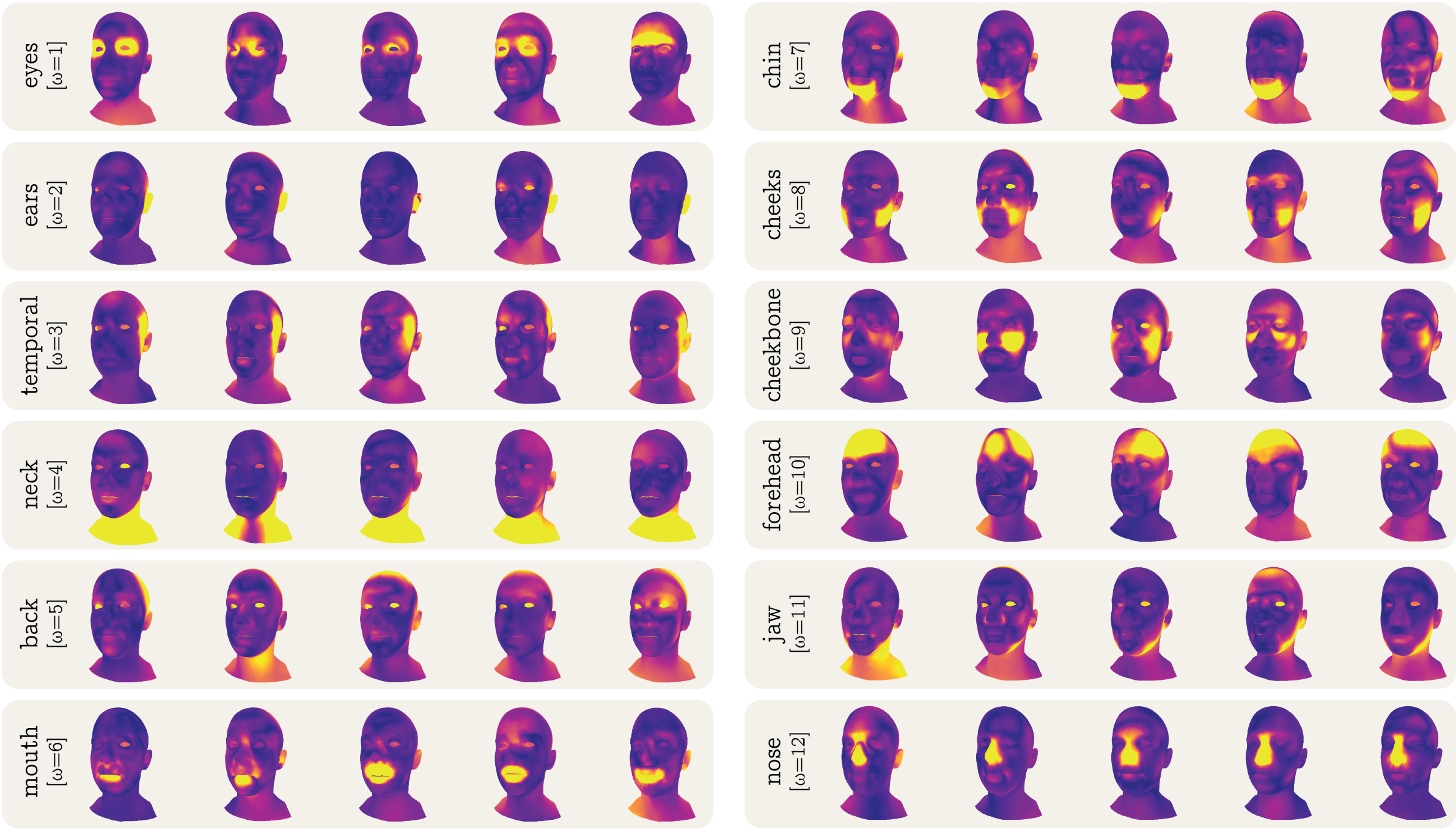}
        \caption{Complete latent traversals of proposed method trained on faces.}
        \label{fig:proposed_latent_exploration}
    \end{figure*}

\begin{figure*}
        \centering
        \includegraphics[width=.9\textwidth]{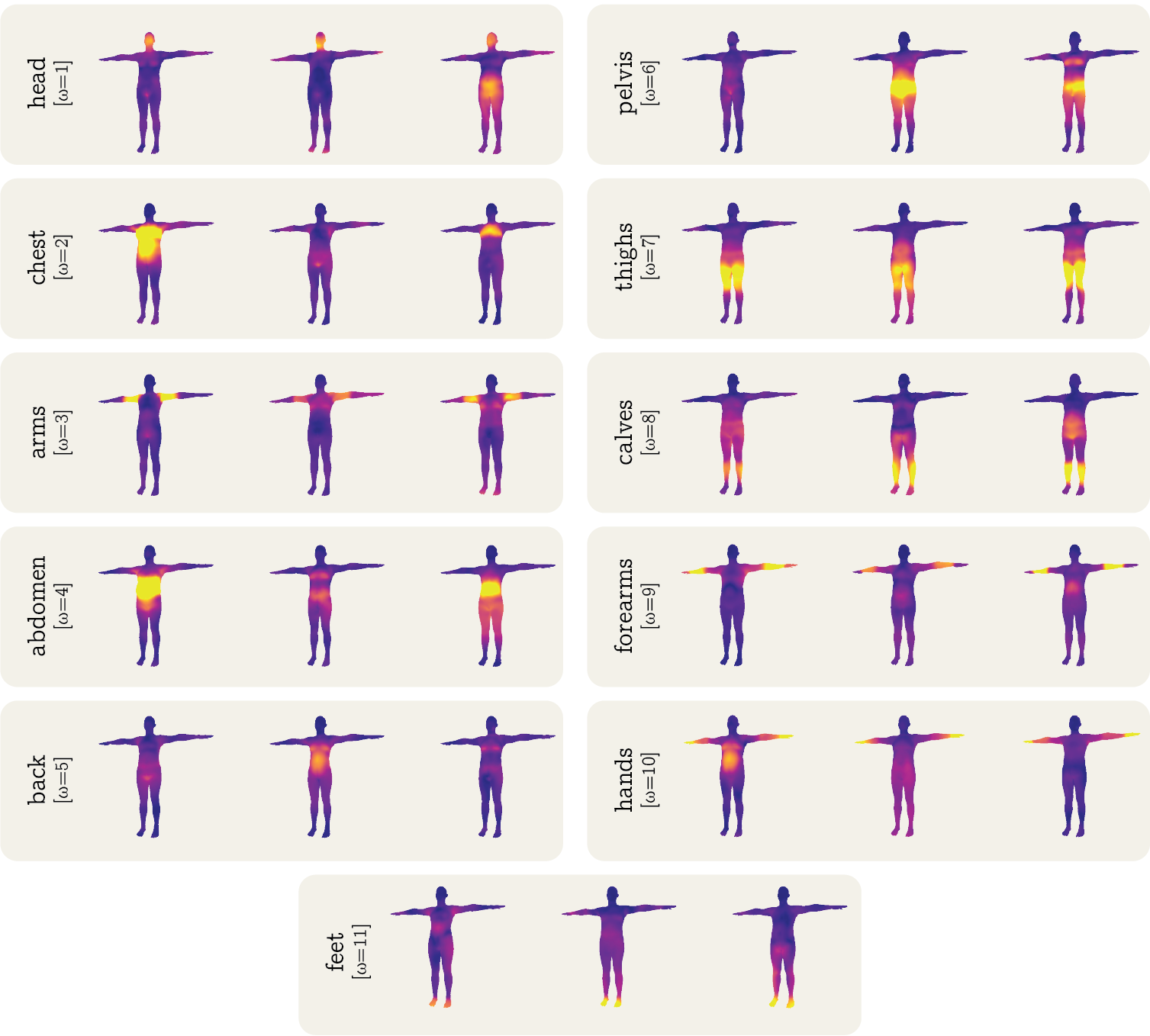}
        \caption{Complete latent traversals of proposed method trained on bodies.}
        \label{fig:proposed_latent_exploration_bodies}
    \end{figure*}

\begin{figure*}
        \centering
        \includegraphics[height=0.95\textheight]{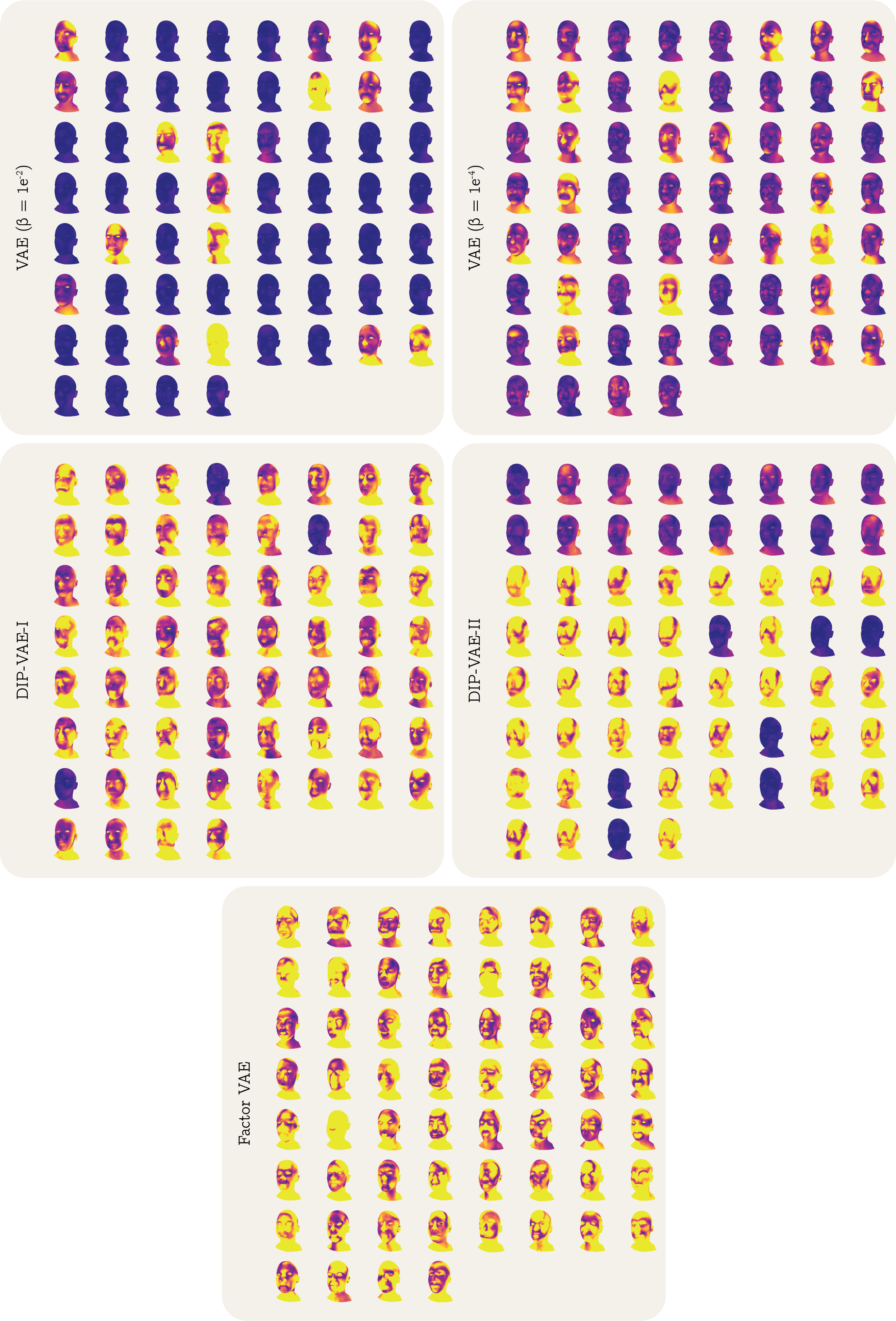}
        \caption{Complete latent traversals for VAE ($\beta = 1e^{-2}$), VAE ($\beta = 1e^{-4}$), DIP-VAE-I, DIP-VAE-II, and Factor VAE.}
        \label{fig:qualitative_comparisons_z_all}
    \end{figure*}

    For each method we report a more comprehensive set of randomly generated samples (Fig.~\ref{fig:qualitative_comparisons_rnd_all}) than those already depicted in Fig.~3.
    Then, we show the full latent disentanglement experiments detailed in Sec.~4. In particular, Fig.~\ref{fig:proposed_latent_exploration} and the \textit{supplementary video} extend Fig.~3 by showing for each $\mathbf{z}^{\omega}$ the effects caused by traversing its latent variables. Similarly, when our method is trained on bodies, Fig.~\ref{fig:proposed_latent_exploration_bodies} extends Fig.~5D. Finally, Fig.~\ref{fig:qualitative_comparisons_z_all} shows the effects of traversing each latent variable of VAE ($\beta = 1e^{-2}$), VAE ($\beta = 1e^{-4}$), DIP-VAE-I, DIP-VAE-II, and Factor VAE. Since these methods do not have a structured latent representation, it is not possible to distinguish different $\mathbf{z}^{\omega}$ like in Fig.~\ref{fig:proposed_latent_exploration}.

\section{Ablation Study}
\begin{figure*}
        \centering
        \includegraphics[width=\textwidth]{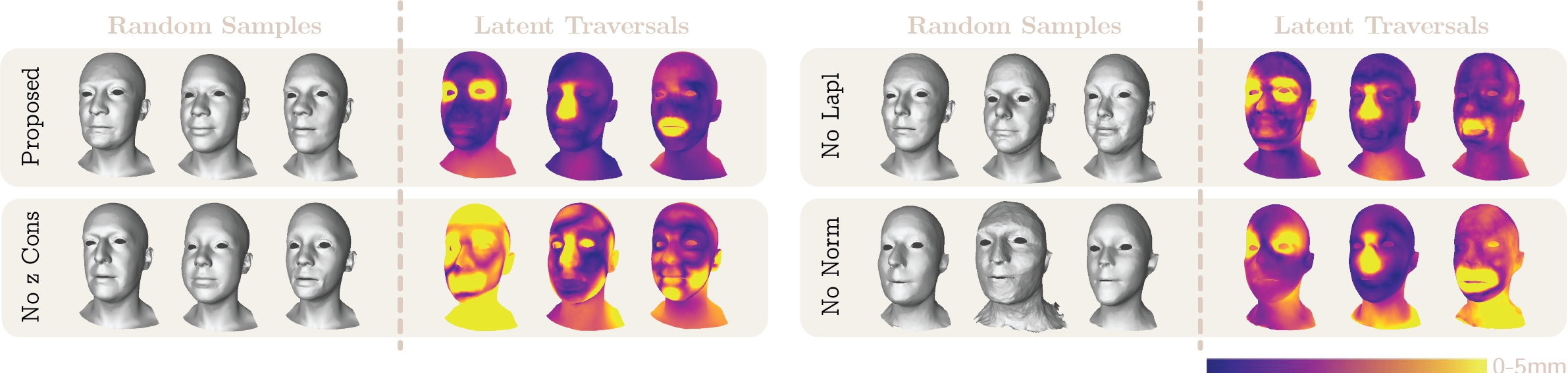}
        \caption{Ablation Study. The proposed method is ablated to examine the effects of the Laplacian regulariser, of the latent consistency loss, and of the instance normalisation. To observe how each of them contributes to the definition of the proposed method we show random samples and vertex-wise distances representing the effects of traversing three randomly selected latent variables.}
        \label{fig:ablation}
    \end{figure*}

    One of the strengths of our method is its intuitiveness and the small number of changes required to convert a VAE into our method. The mini-batch feature swapping and the latent consistency loss are indeed the only changes required, but we decide to analyse also two important components that characterise our implementation of the VAE: the Laplacian regulariser used in the loss function and the instance normalisation. The ablation study, whose results are depicted in Fig.~\ref{fig:ablation}, is performed re-training the proposed model with the necessary modifications. When $\kappa$ is set to $0$, $\mathcal{L}_{c}$ is ignored (see Eq.~3). Despite this appears to be equivalent to the VAE, note that in this case the mini-batch feature swapping is performed. Observing the latent perturbations in Fig.~\ref{fig:ablation}~(No z Cons) we see the importance of the latent consistency loss. As expected, curating only the mini-batching does not allow to obtain a structured and disentangled latent space. Observing the random samples depicted in Fig.~\ref{fig:ablation}~(No Lapl) and obtained from the proposed method re-trained with $\alpha=0$, we notice that the contributions of $\mathcal{L}_{L}$ are more subtle. Nevertheless when this term is removed we notice a more irregular surface as well as some surface discontinuity (e.g. top part of the head in the first sample or neck of the second sample). Finally, from Fig.~\ref{fig:ablation}~(No Norm) we observe the importance of the normalisation, which helps the generation of realistic faces.
    
 \section{Generalisation Capabilities}
\begin{figure*}
        \centering
        \includegraphics[width=0.8\textwidth]{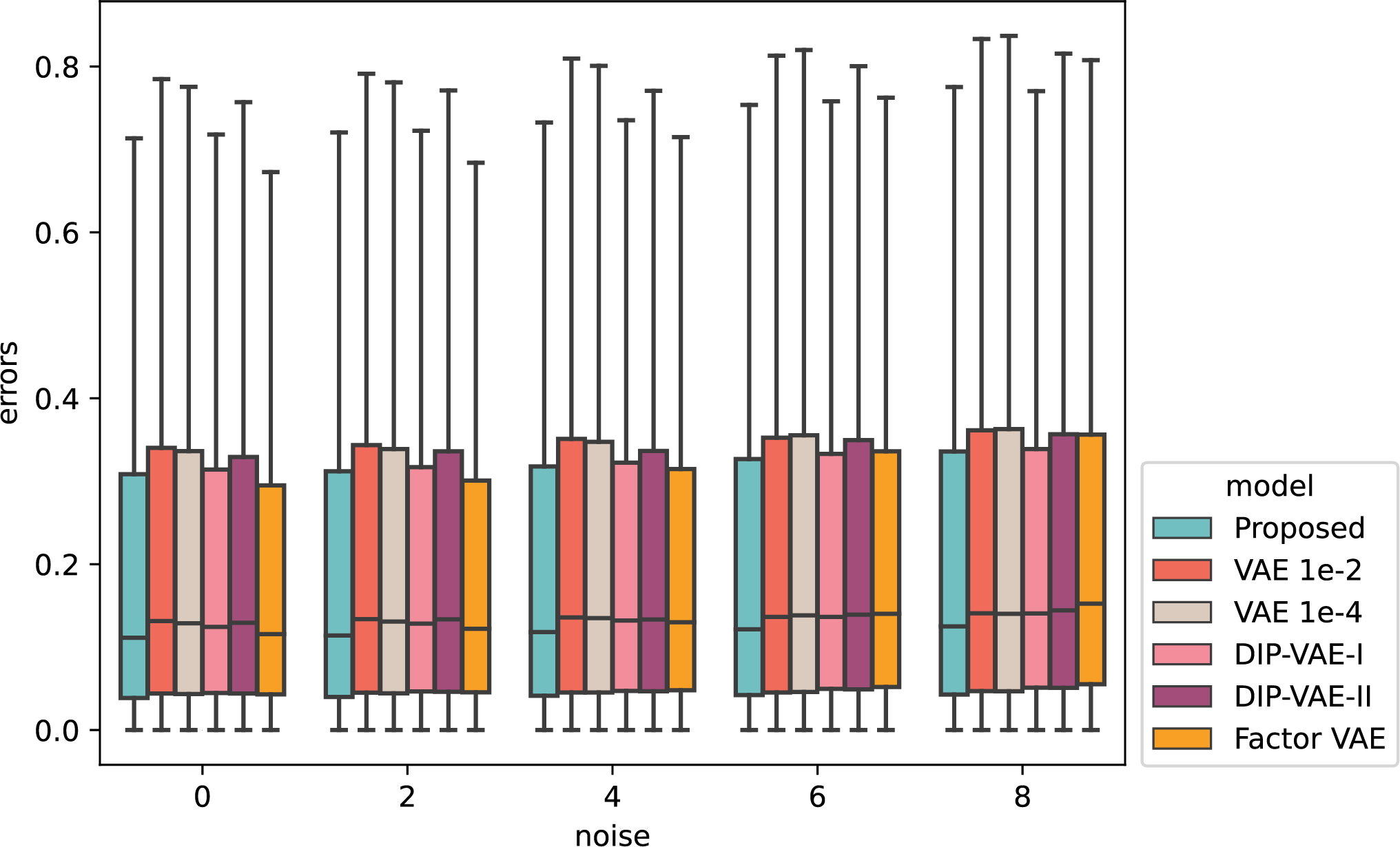}
        \caption{Generalisation capabilities evaluated by fitting CoMA subjects in neutral expressions with the proposed method as well as VAE ($\beta = 1e^{-2}$), VAE ($\beta = 1e^{-4}$), DIP-VAE-I, DIP-VAE-II, and Factor VAE.}
        \label{fig:fit}
    \end{figure*}

    We evaluate the ability of our model to generate meshes outside the training data distribution by fitting all the CoMA subjects in their neutral expressions. Starting from the mean latent representation, we iteratively generate new meshes. For the first $80$ iterations we optimise $\mathbf{z}$ with a mean squared error over $24$ manually selected facial landmarks, for the remaining $170$ iterations we use a Chamfer distance between the vertices generated with our model and those of the target mesh from CoMA. Also for this experiment we use the \textsc{Adam} optimiser setting the learning rate to $lr=5e^{-3}$. Errors are computed as per-vertex distances between each generated vertex and the closest vertex of the target mesh. To evaluate the robustness to noise we repeat the experiment perturbing the target vertices of meshes from CoMA with different amounts of random noise. When noise is applied, errors are computed with respect to the target without noise. As we could expect, in Fig.~\ref{fig:fit}, we show that errors increase linearly with the amount of noise applied for all methods. Nevertheless, errors remain low, thus proving good generalisation capabilities. In addition, our method is the one with the lowest errors despite all methods were trained on the same dataset.
    
\clearpage
\section{Societal Impact}
    Our work focuses on the generation of 3D shapes of bodies and faces, but shapes without textures and materials are far from being realistic. For this reason, we believe that our work does not raise disinformation or immediate security concerns. Nevertheless, it still involves sensitive human data and solutions to disentanglement could potentially find future applications to face image manipulation. 
    
    Finally, considering that the limited size of our model does not require long trainings (see Implementation Details in Sec.~4), the proposed method does not cause significant environmental impacts.